\documentclass[conference]{IEEEtran}
\IEEEoverridecommandlockouts
\usepackage{cite}
\usepackage{amsmath,amssymb,amsfonts}
\usepackage{graphicx}
\usepackage{booktabs}
\usepackage{multirow}
\usepackage{xcolor}
\usepackage{url}
\usepackage[hidelinks]{hyperref}
\usepackage{balance}

\usepackage{caption}
\newcommand{\affilmark}[1]{\textsuperscript{#1}}
\newcommand{\advisemark}{\textsuperscript{\dag}}

\begin{document}

\title{Try Once, Then Optimal:\\ De-Redundified Procedure Memory for\\ Cross-Episode Exploration Amortization}

\author{
Haizhou Ge\affilmark{1}, Haochen Ouyang\affilmark{3}, Zhixing Chen\affilmark{1}, Yufei Jia\affilmark{1}, Yue Li\affilmark{2},\\
Lu Shi\affilmark{1}\advisemark, Lei Han\affilmark{2}\advisemark, Guyue Zhou\affilmark{1}\advisemark, Ruqi Huang\affilmark{1}\advisemark
\thanks{\advisemark Corresponding author.}
\thanks{\affilmark{1}Tsinghua University.
        {\tt\small \{ghz23,jyf23,chenzx24\}@mails.
        tsinghua.edu.cn,
        ruqihuang@sz.tsinghua.edu.cn,
        \{shilu,zhouguyue\}@air.tsinghua.edu.cn}}
\thanks{\affilmark{2}DISCOVER Robotics.
        {\tt\small \{lue,leo\}@discover-robotics.com}}
\thanks{\affilmark{3}Northeast Agricultural University.
        {\tt\small a07230190@neau.edu.cn}}
}

\maketitle
\begin{abstract}
Manipulating objects with hidden internal state, such as a latched microwave, forces a robot to probe before it can act. Yet a robot that has solved an instance once re-runs the same probes whenever it encounters that instance again, because existing cross-episode memories target task success and organize reuse around states, not the object or the cost of re-exploring it. We present \textbf{Instance-Oriented Memory (IOM)}, an object-centric framework that amortizes this exploration: from a single encounter that uncovers the hidden state, whether or not it succeeds, IOM records a short procedure for manipulating that instance, keys it on the object's identifiable features, and injects it as a soft bias on a procedure-conditioned policy. A later encounter recognizes the object and recalls its procedure instead of re-exploring. We instantiate this distillation with an off-the-shelf vision--language model (VLM) that parses each encounter into the procedure without task-specific training. Across four articulated-object tasks, two in simulation (microwave, door) and two on a real robot (bottle, cabinet), an oracle procedure memory cuts manipulation operations by \textbf{16--30\%} over re-exploration at non-regressing success, and the VLM instantiation recovers \textbf{69--88\%} of that saving out of the box. Because the procedure is a soft bias on a feedback-driven policy, an incorrect memory is recovered from rather than obeyed: success holds even when a retrieved procedure is wrong, as for $\approx$12\% of door instances. Across all tasks the benefit is purely one of \textbf{efficiency}: success never regresses, and on the real robot even improves. Code will be released upon acceptance.
\end{abstract}


\section{Introduction}
\label{sec:intro}
\begin{figure*}[t]
    \centering
    \includegraphics[width=0.88\textwidth]{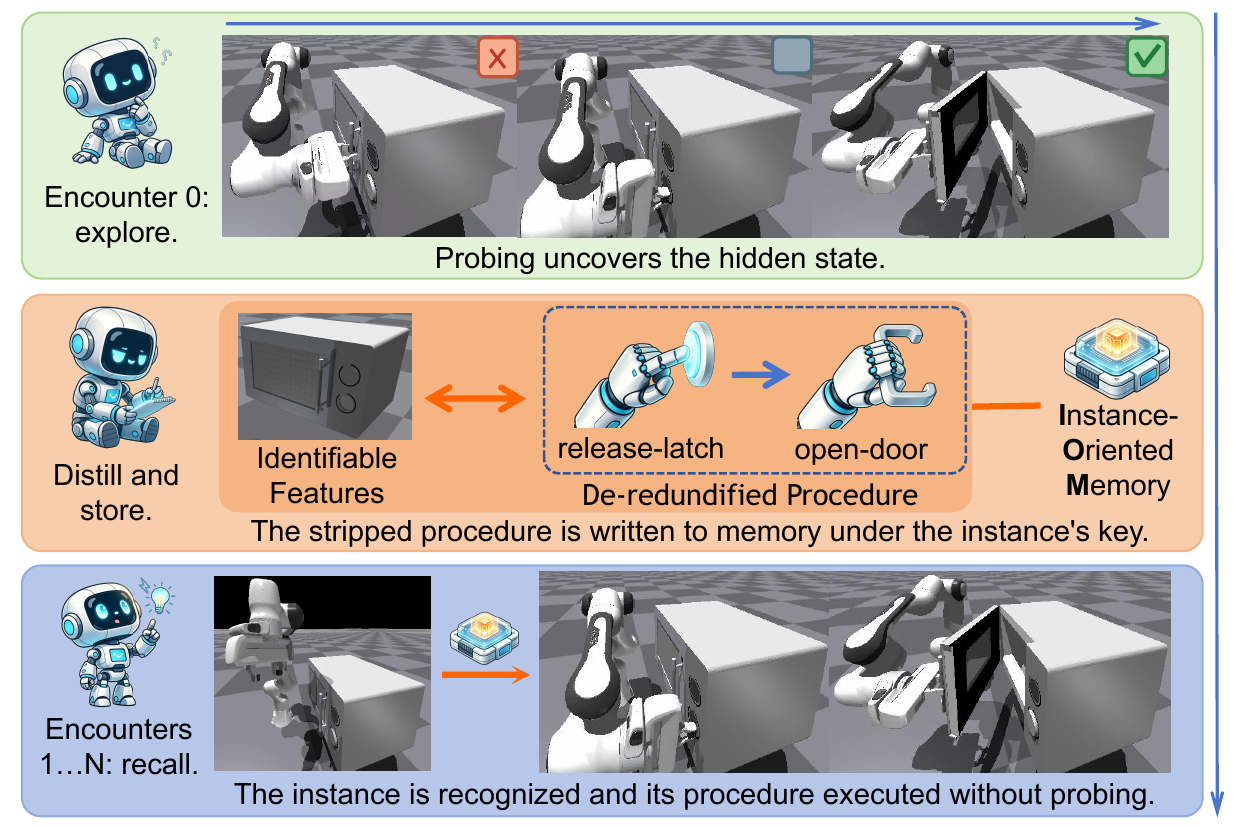}
    \caption{\textbf{Instance-Oriented Memory (IOM).} An object with hidden internal state (here a latched microwave) must be probed before it can be manipulated, and that probe is pure overhead on every later encounter. IOM removes it. \textbf{Top}: the first encounter probes, which uncovers the hidden state whether or not the attempt succeeds. \textbf{Middle}: distillation strips the redundant probe into a short \emph{de-redundified procedure}, written to an instance-keyed memory under the object's identifiable features $\phi(x)$. \textbf{Bottom}: a later encounter recognizes the instance, recalls its procedure, and executes it without probing. The procedure enters the policy as a soft bias, so an incorrect memory is recovered from rather than obeyed: the saving is one of efficiency, with operations falling \textbf{16--30\%} on repeat encounters at non-regressing success.}
    \label{fig:teaser}
\end{figure*}

Many everyday objects carry a hidden internal state that a robot cannot see before touching them, such as a microwave whose latch may be engaged. Acting on such an object requires probing, the robot must interact to reveal the state before it can choose the right manipulation. Prior knowledge narrows this but does not always remove it: a manual gives the generic procedure, not whether this particular latch is currently engaged, and a status cue such as a lock indicator light resolves that only when present, working, and recognized by the policy. So probing is unavoidable the first time, yet pure overhead every time after. A robot that has already opened a particular microwave should not, each morning, cautiously release a latch it has already learned is free.

Existing cross-episode memories pursue a different goal. To improve task success or robustness, they give a policy a memory of past successful experience and retrieve it, by the current observation, as an action snippet to execute~\cite{kwon2025rtcache} or an action prior that steers the policy~\cite{zhao2026retrievethensteer}. Organized around states and experiences rather than around the object, they are not built to recognize a recurring instance and skip the exploration it no longer needs. A separate line resolves an instance's hidden state by trial and error within a single deployment~\cite{wang2025adamanip, wang2022adaafford}, but discards that knowledge once the episode ends.

We take an object-centric view: a robot should remember the objects it manipulates and how to manipulate them. We present \textbf{Instance-Oriented Memory (IOM)} (Fig.~\ref{fig:teaser}), a modular framework that amortizes exploration across repeated encounters with the same object instance. From a single encounter that uncovers the hidden state, whether or not it succeeds, IOM distills a short procedure for manipulating that instance, keys it on the object's identifiable features, and injects it as a soft bias on a procedure-conditioned policy. The first encounter explores; every later one recognizes the object, recalls its procedure, and executes it directly instead of re-exploring. Because the procedure enters only as a soft bias, an unfamiliar object or a wrong recollection falls back to exploration rather than to failure. Targeting efficiency, the axis these success-oriented memories leave open, IOM composes with them: it layers onto an existing method to cut redundant operations, leaving that method's success mechanism untouched.

We make the following contributions:
\begin{itemize}
\item \textbf{The IOM framework.} We cast cross-episode exploration amortization as an object-centric memory problem and realize it as three interchangeable modules (procedure distillation, instance-keyed memory, and a procedure-conditioned policy), each a general signature that admits many realizations. Across four articulated-object tasks, two simulated and two on a real robot, an oracle procedure memory cuts manipulation operations by 16--30\% over re-exploration on repeat encounters at non-regressing success, isolating the value of the memory itself.
\item \textbf{A training-free instantiation.} We realize distillation and memory with an off-the-shelf vision-language model (VLM) that parses each encounter into a procedure and recalls it by appearance, with no task-specific training. This instantiation recovers 69--88\% of the oracle saving out of the box, reaching most of the framework's benefit with no learned distillation or memory.
\end{itemize}

\section{Related Work}
\label{sec:related}

\subsection{Cross-episode memory for manipulation policies}
Closest to our setting, a policy is given a memory of past successes to reuse at test time, with the goal of improving success or robustness. RT-Cache~\cite{kwon2025rtcache} retrieves and executes action snippets from a trajectory store; Retrieve-then-Steer~\cite{zhao2026retrievethensteer} aggregates successful segments into an action prior that steers the sampler; MAP-VLA~\cite{li2025mapvla} retrieves stage-indexed soft prompts; others personalize a policy to an instance from reference images~\cite{lee2025vap} or retrieve past keyframes for a low-level policy~\cite{sridhar2026memer}, echoing earlier instance-level model reuse~\cite{lu2021objectreuse}. These memories are keyed and retrieved by the current observation and serve task success, which their raw action units do well. IOM differs in objective and organization rather than in artifact quality: it is object-centric and targets operation efficiency, keying memory on the object instance and recording, per object, the manipulation procedure that lets a later encounter skip exploration. Retrieve-then-Steer is the closest in mechanism, an online success memory on a feedback-driven policy, but it keys on the current state to raise success, whereas IOM keys on the object to remove repeated exploration.

\subsection{Adaptive manipulation under hidden state}
A separate body of work resolves an object's hidden internal state by interacting with it. AdaManip~\cite{wang2025adamanip} discovers the right sequence (probe the lock; if locked, release it, then open) by trial and error per attempt; AdaAfford~\cite{wang2022adaafford} adapts an affordance prior to an instance from a few test-time interactions; related work grounds the hidden state with foundation-model reasoning~\cite{zhang2025adaptiveonthefly} or with interaction-driven affordance trajectories~\cite{wu2022vatmart}, and benchmarks the resulting multi-stage, non-Markovian behavior~\cite{wang2026vqmem}. These methods solve the within-episode discovery that IOM also relies on, but they discard it when the episode ends. IOM is the memory layer that persists the per-instance procedure and reuses it across encounters.

\subsection{Retrieval-augmented and in-context imitation}
Retrieval has also been used for data efficiency, retrieving sub-trajectories or skills to augment policy training~\cite{memmel2025strap, nasiriany2022sailor}, and in-context imitation conditions a policy on demonstrations supplied as a prompt without weight updates~\cite{dipalo2024kat, fu2024icrt, vosylius2025instant}. In both, the retrieved or prompted unit is raw demonstration data, supplied offline or by a human. IOM's conditioning is instead a procedure the system abstracts from its own encounter, keyed on the instance and written online.

\subsection{Memory and skill libraries in agents and robots}
The store-abstract-retrieve loop is well established for language agents, which reflect on trials into reusable text and recall it later~\cite{shinn2023reflexion, zhao2024expel, park2023generative, allard2026erl, pink2025episodic}, and for robots that distill corrections or skills into a growing library~\cite{zha2023droc, tziafas2024lrll, sarch2023helper, wang2023voyager, liu2023reflect}. IOM shares this loop but operates at the low-level control interface: the abstracted artifact conditions a visuomotor policy, which retains the ability to re-explore when the memory is wrong. Finally, ``memory'' in many recent vision-language-action models is \emph{intra}-episode temporal context for partial observability~\cite{shi2025memoryvla, yang2026eventvla, torne2026mem}, a different problem from learn-once-reuse-across-encounters, and self-improving robots reach comparable gains but require many practice trials and weight updates~\cite{bousmalis2024robocat, sharma2023medal}, whereas IOM amortizes from a single encounter. The benchmarks that probe robot memory share this scope, scoring intra-episode recall or final task success~\cite{lei2026robomemarena, wang2026vqmem}; to the best of our knowledge none measures the operation cost of re-encountering a known instance. Absent an object-centric efficiency benchmark, we build our evaluation on the hidden-state articulated tasks of AdaManip~\cite{wang2025adamanip} and extend them to a real robot.

\section{Method}
\label{sec:method}
\begin{figure*}[t]
    \centering
    \includegraphics[width=0.95\textwidth]{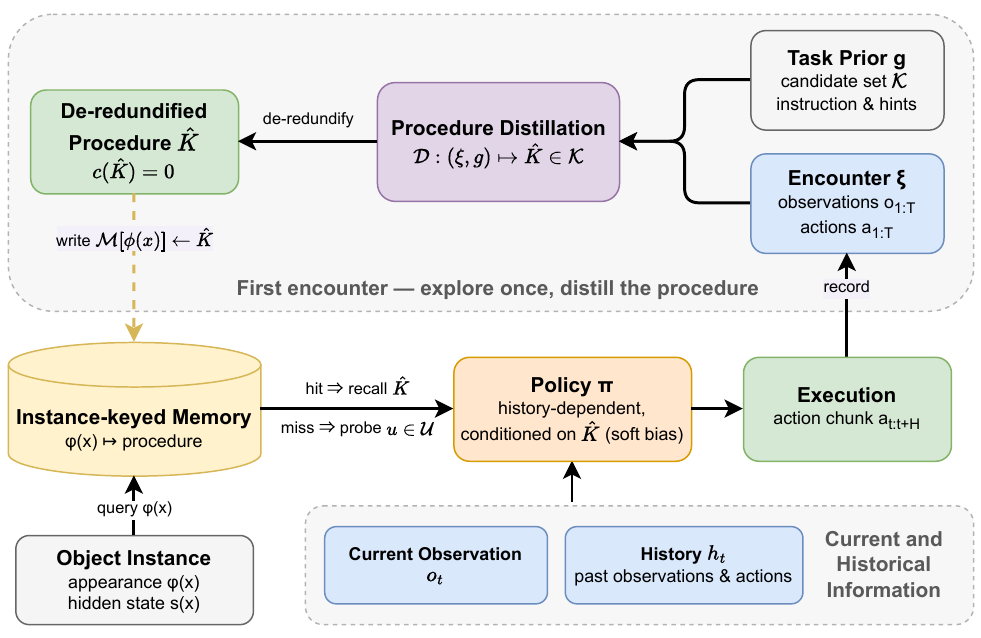}
    \caption{\textbf{The Instance-Oriented Memory (IOM) framework.} IOM amortizes exploration across repeated encounters with the same object instance through three modules: procedure distillation, instance-keyed memory, and a procedure-conditioned policy. \textbf{Bottom (shared spine)}: an object instance is keyed by its identifiable features $\phi(x)$ into the \emph{instance-keyed memory}; a hit recalls the stored procedure $\hat{K}$, a miss returns an uncovering probe $u \in \mathcal{U}$. The history-dependent \emph{policy} $\pi$ consumes this together with the current observation $o_t$ and history $h_t$ (bottom-right) and rolls out an action chunk. \textbf{Top (first encounter)}: on the first encounter the rollout is recorded as an \emph{encounter} $\xi$, and \emph{procedure distillation} $\mathcal{D}\!:(\xi, g)\mapsto\hat{K}\in\mathcal{K}$ strips the redundant probe (using the shared task prior $g$) into a minimal de-redundified procedure $\hat{K}$ with $c(\hat{K})=0$, which is then written back to memory under the instance key. Every later encounter recalls $\hat{K}$ and executes it directly instead of re-exploring. The procedure is injected as a \emph{soft bias} on the policy, so an incorrect memory is recovered from rather than obeyed.}
    \label{fig:method}
\end{figure*}

We present Instance-Oriented Memory (IOM, Fig.~\ref{fig:method}), a modular memory framework that amortizes exploration across repeated \emph{encounters} with the same object instance, where each encounter is one episode in which the robot attempts the task. We first formalize the setting and objective (\S\ref{sec:problem}) and then specify the framework's three modules, each given first in a general form that fixes its signature and then in the instantiation we evaluate (\S\ref{sec:mod-distill} through \S\ref{sec:mod-policy}), before turning to how the policy is trained (\S\ref{sec:training}) and how the candidate procedures the modules choose among are sourced (\S\ref{sec:sourcing}).

\subsection{Problem formulation}
\label{sec:problem}
Consider a class of objects $\mathcal{C}$ and a manipulation task $\tau$ defined on it, for instance opening a microwave. The task is carried out by sequences of atomic operations. Let $\mathcal{P}$ be the set of all sequences that complete $\tau$. Because an object's internal state is hidden, resolving it requires interaction, and completing $\tau$ thus generally involves trial and error; the repeated sub-attempts this leaves in a sequence are its \emph{retries}. The \emph{no-retry} sequences $\mathcal{R} \subseteq \mathcal{P}$ are those without any.

An object \emph{instance} $x \in \mathcal{C}$ carries a hidden internal state $s(x)$, possibly comprising several latent variables such as latches or locks, that is not observable before interaction. Relative to this state, a no-retry sequence is \emph{non-redundant} for $x$ when no operation can be dropped from it while still completing $\tau$ under $s(x)$, which makes non-redundancy an instance-level property. For an already-unlocked microwave, \texttt{[release-latch, open-door]} carries no retry yet is redundant, because the release is unnecessary, whereas \texttt{[open-door]} is non-redundant. Across the possible hidden states, the task admits a finite set of \emph{de-redundified procedures} $\mathcal{K} = \{K_1, \dots, K_m\} \subseteq \mathcal{R}$, and the non-redundant procedures for a given $x$ form a subset
\begin{equation}
\mathcal{K}^\star(x) = \kappa\bigl(s(x)\bigr) \subseteq \mathcal{K},
\label{eq:kstar}
\end{equation}
with $\kappa$ a map from internal state to procedures; this subset is typically a singleton, with several elements for example when $\tau$ admits operations whose order is immaterial.

Each encounter executes one sequence $p \in \mathcal{P}$, with cost $c(p)$ the number of redundant operations it contains, those droppable while still completing $\tau$ under $s(x)$, so $c(p) = 0$ exactly when $p$ is non-redundant. Acting optimally requires knowing $s(x)$, which only interaction reveals: a sequence is \emph{uncovering} if attempting it determines $s(x)$ in full, and we write $\mathcal{U}$ for the set of such sequences. For the microwave, \texttt{[open-door]} is uncovering, since the door yielding or resisting identifies the latch, whereas \texttt{[release-latch, open-door]} is not, succeeding either way and revealing nothing; the first encounter therefore probes with a sequence from $\mathcal{U}$.

Writing $p_i$ for the $i$-th encounter's sequence ($i = 0, 1, 2, \dots$), a memoryless agent re-probes at every encounter and repeatedly incurs redundancy. IOM instead discovers $s(x)$ once, from the uncovering first encounter, and reuses it: every later encounter executes a non-redundant procedure $K \in \mathcal{K}^\star(x)$, so $c(p_i) = 0$ for all $i > 0$. The first encounter's exploration is thereby amortized over the repeats.

An instance's hidden state may \emph{reset} between encounters or \emph{persist} across them. Under reset, the state returns unchanged each time, so a cabinet re-locked between visits requires \texttt{[unlock, open-door]} on every encounter. Under persistence, the operations of one encounter reshape the next: a cabinet left unlocked is thereafter opened by \texttt{[open-door]} alone, and the optimal procedure becomes history-dependent. IOM provides the same value in both settings, so our experiments use only the reset setting; letting the robot instead determine an instance's current hidden state on its own, for example through a long-term memory of past interactions, is left to future work.

\subsection{Procedure distillation}
\label{sec:mod-distill}
From the first encounter's possibly redundant sequence, IOM must recover the compact procedure that later encounters will reuse. The distillation module does so, mapping an encounter $\xi$ and the task prior $g$ to a de-redundified procedure,
\begin{equation}
\mathcal{D}: (\xi, g) \mapsto \hat{K} \in \mathcal{K},
\label{eq:distill}
\end{equation}
where $g$ is shared across all instances of $\mathcal{C}$ and supplies a vague instruction, the candidate set $\mathcal{K}$, and optional hints. Once $\xi$ has revealed $s(x)$, the module strips the redundant operations from the executed sequence and returns $\hat K$, ideally with $c(\hat K) = 0$. Because the non-redundant set depends only on $s(x)$ (Eq.~\ref{eq:kstar}), distillation is governed by revelation rather than completion: an encounter that fails the task but still reveals $s(x)$ yields the same $\hat K$ as a success would. Any map of this form fills the module, whether a deterministic extractor, a learned encoder, or a vision-language model.

We instantiate $\mathcal{D}$ with a pretrained vision-language model (VLM). The encounter $\xi$ is presented as an RGB observation video $o_{1:T}$ and the executed action trajectory $a_{1:T}$, and $g$ as a language prompt: a vague task instruction (for instance, \emph{open the microwave}), the candidate set $\mathcal{K}$, and hints on what to produce and how to interpret the encounter. The VLM returns $\hat K \in \mathcal{K}$ with no task-specific post-training.

\subsection{Instance-keyed memory}
\label{sec:mod-memory}
A distilled procedure is useful only if the same instance can be recognized when it recurs and its procedure recalled. The memory module provides this, a partial map $\mathcal{M}$ from an instance's identifiable features $\phi(x)$, such as its appearance or a model identifier, to a stored procedure. On revelation it writes $\mathcal{M}[\phi(x)] \leftarrow \hat K$, and on a later encounter it returns
\begin{equation}
\mathcal{M}\bigl(\phi(x)\bigr) =
\begin{cases}
\hat K, & \text{if } x \text{ has been written},\\
u, & \text{otherwise},
\end{cases}
\label{eq:memory}
\end{equation}
where the second case, $u \in \mathcal{U}$, is an uncovering probe that reveals $s(x)$ and triggers the write that serves every later encounter. Any keyed associative store, parametric or not, fills the module. Separate from the policy, the memory can be shared across tasks: separately trained policies, single-task or not, draw on the same store, each recalling the procedure for the instance it faces.

We realize $\mathcal{M}$ with a pretrained vision-language model. To write, the model associates the object's first frame with the procedure $\hat K$. On a later encounter it is shown the new first frame and asked whether the object it must manipulate is one it has already seen; if so it returns the associated procedure, and otherwise an empty string, so that an unfamiliar object is probed with a sequence from $\mathcal{U}$, as at a first encounter.

The module stores whatever distillation returns, so a procedure that misreads the hidden state is recalled at every later encounter, reintroducing the redundant operations the memory exists to remove; the soft bias keeps these recalls successful (\S\ref{sec:training}), but the instance loses its amortization, and the map as specified cannot repair the entry. A \emph{meta-memory} that monitors the outcomes of its own recalls~\cite{liang2026metamemory} would close this gap: a faulty entry betrays itself through the retries it provokes, and those retries reveal $s(x)$ anew, so distillation can re-derive the correct procedure and overwrite the stored one. We leave this self-correction to future work.

\subsection{Procedure-conditioned policy}
\label{sec:mod-policy}
To operate within IOM, the policy must meet two requirements. First, it must admit the retrieved $K$ as a conditioning input alongside its observations. Second, because the action distribution in these tasks is often multi-modal, it must be history-dependent rather than memoryless, conditioning on past observations and actions rather than the current observation alone. It therefore maps the current observation $o_t$, the history $h_t$, the procedure $K$, and any further conditioning $z$ to an action chunk~\cite{zhao2023act} of horizon $H$,
\begin{equation}
a_{t:t+H} \sim \pi\bigl(\cdot \mid o_t,\, h_t,\, K,\, z\bigr), \qquad K \in \mathcal{K} \cup \mathcal{U}.
\label{eq:policy}
\end{equation}
Such conditioning might be a multi-task policy's language instruction, and is absent when the policy uses none; the framework is indifferent to it, as $K$ enters identically regardless, so any history-dependent policy that can be conditioned on the retrieved procedure fills the module. Separating $K$ from $z$ is conceptual rather than architectural: when both are natural language, $K$ may simply be appended to $z$ and the two consumed through a single channel.

We adapt the policy of AdaManip~\cite{wang2025adamanip}, replacing its diffusion policy with flow matching and adding a language-conditioning channel that carries $K$; the setup is single-task, so $z$ is empty. The prompt $K$ is embedded by a pretrained text encoder~\cite{wang2023m3e}, projected to a lower dimension, and concatenated after the flattened observation features along the feature dimension to form the conditioning vector.

\subsection{Training for soft-bias conditioning}
\label{sec:training}
For the sequence $p$ executed in an encounter $\xi$, we collect into a set $\mathcal{K}^\circ(p)$ the conditions that open with $p$'s operations before its first retry but are otherwise unconstrained, and draw $K$ from it at random during training. On a locked microwave, a sequence $p$ that opens with a failed \texttt{[open-door]} and succeeds only on retry with \texttt{[release-latch, open-door]} yields $\mathcal{K}^\circ(p)$, the conditions opening with that failed \texttt{[open-door]} rather than the eventual fix. Table~\ref{tab:kcircle} gives such an encounter and its $\mathcal{K}^\circ(p)$ for each task. The policy thus learns to follow $K$ at the opening yet rely on feedback beyond it: it carries out a correct $K \in \mathcal{K}^\star(x)$ and recovers through retries when execution diverges from any other $K$. We measure this recovery in \S\ref{sec:results}, where success does not regress even when the supplied $K$ is wrong.

We expect this training step to grow unnecessary as embodied foundation models mature: a generalist policy with robust everyday-manipulation skills and reliable instruction following should apply this soft bias natively, deferring to interaction feedback when procedure and feedback disagree rather than carrying the procedure out regardless. Large-scale pretraining may already instill this behavior implicitly, much as it confers a general capacity for error detection and recovery, in which case $K$ enters such a policy as an ordinary instruction with no dedicated training stage. Recent work has diagnosed whether vision-language policies rely on language grounding or visual shortcuts when the two conflict~\cite{lin2026la4vla}, but to the best of our knowledge no existing benchmark isolates whether a policy equipped with procedure memory defers to feedback when a retrieved procedure is wrong. We leave the measurement of this native deferral capacity to future work.

\subsection{Sourcing the procedure sets}
\label{sec:sourcing}
The modules draw on three sets, fixed once before deployment: the candidate procedures $\mathcal{K}$, the uncovering probes $\mathcal{U}$, and the training conditions $\mathcal{K}^\circ(p)$. Only $\mathcal{K}$ must be obtained from outside the task; the other two generally follow from it. A sourcing map returns the candidates from the object class and the task,
\begin{equation}
\mathcal{K} = \Sigma(\mathcal{C}, \tau),
\label{eq:source}
\end{equation}
after which the uncovering probes are the sequences whose outcomes separate these candidates by hidden state, $\mathcal{U} = \nu(\mathcal{K})$, and the training conditions $\mathcal{K}^\circ(p)$ are read from $\mathcal{K}$ and the executed sequence $p$ as in \S\ref{sec:training}.

Any map that returns the class's candidate procedures fills $\Sigma$: an LLM queried for the procedures its pretraining already covers, or, for a specific market model, an agent that retrieves the product manual online and reads the procedures from it. Deriving the other two sets is likewise undemanding: recovering $\mathcal{U}$ from $\mathcal{K}$ through $\nu$ is within reach of an LLM's reasoning, while $\mathcal{K}^\circ(p)$ amounts to matching $p$'s pre-retry opening against $\mathcal{K}$, a deterministic filter that a short script or the LLM applies equally well.

We construct all three sets by hand from each task's known structure. Fixing them exactly isolates the modules under study from sourcing error, so that the differences we report reflect the stored procedure rather than the quality of set construction.

\begin{table*}[t]
\centering
\setlength{\tabcolsep}{5pt}
\caption{Procedure sets for the four evaluation tasks. Each task has a binary hidden state; $\mathcal{K}$ holds the two non-redundant procedures, $\mathcal{U}$ the uncovering probe, and $\mathcal{P}\setminus\mathcal{R}$ a retry-containing sequence, where $\setminus$ denotes set difference. Operation names are abbreviated (\texttt{open}~=~open-door, \texttt{pull}~=~pull-drawer, \texttt{release}~=~release-latch, \texttt{unscrew}~=~unscrew-cap, \texttt{lift}~=~lift-cap, \texttt{cw}~=~turn-clockwise, \texttt{ccw}~=~turn-counterclockwise). $^\dagger$For \emph{door} the two states are symmetric, so the probe may start in either direction; the \texttt{cw} case is shown.}
\label{tab:sets}
\begin{tabular}{llll}
\toprule
Task & $\mathcal{K}$ (non-redundant) & $\mathcal{U}$ (uncovering probe) & $\mathcal{P}\setminus\mathcal{R}$ (with retry) \\
\midrule
microwave & \texttt{[open]}, \texttt{[release, open]} & \texttt{[open]} & \texttt{[open, release, open]} \\
door & \texttt{[cw, open]}, \texttt{[ccw, open]} & \texttt{[cw, open]}$^\dagger$ & \texttt{[cw, ccw, open]}$^\dagger$ \\
bottle & \texttt{[lift]}, \texttt{[unscrew, lift]} & \texttt{[lift]} & \texttt{[lift, unscrew, lift]} \\
cabinet & \texttt{[pull]}, \texttt{[release, pull]} & \texttt{[pull]} & \texttt{[pull, release, pull]} \\
\bottomrule
\end{tabular}
\end{table*}

\begin{table}[h]
\centering
\setlength{\tabcolsep}{3pt}
\caption{Example encounters and the training conditioning sets $\mathcal{K}^\circ(p)$ they yield. Each encounter executes a sequence $p$ that fails initially and succeeds on retry; $\mathcal{K}^\circ(p)$ comprises all conditions in $\mathcal{K}$ opening with $p$'s pre-retry prefix, training the policy to follow that opening yet recover when it diverges.}
\label{tab:kcircle}
\begin{tabular}{lp{4.2cm}l}
\toprule
Task & Encounter sequence $p$ & $\mathcal{K}^\circ(p)$ \\
\midrule
microwave & \texttt{[open, release, open]} & \texttt{[open]} \\
door & \texttt{[cw, ccw, open]} & \texttt{[cw, open]} \\
bottle & \texttt{[lift, unscrew, lift]} & \texttt{[lift]} \\
cabinet & \texttt{[pull, release, pull]} & \texttt{[pull]} \\
\bottomrule
\end{tabular}
\end{table}

\section{Experiments}
\label{sec:experiments}
Our evaluation reports IOM's overall efficiency and ablates its two contributions, the framework and its training-free instantiation, then tests robustness to a wrong memory:
\begin{enumerate}
\item \textbf{Framework} (memory vs.\ none): does storing and reusing a de-redundified procedure cut operations on repeat encounters rather than re-exploring from scratch?
\item \textbf{Instantiation} (oracle vs.\ off-the-shelf): does the training-free vision--language realization capture that reduction out of the box?
\item \textbf{Robustness}: is an incorrect or stale procedure recovered from rather than followed into failure?
\end{enumerate}
We evaluate in simulation on two AdaManip~\cite{wang2025adamanip} articulated-object tasks (\emph{microwave}, \emph{door}) over $90$ instances ($50$ and $40$) with $5$ episodes each, and on two real-robot tasks (\emph{bottle}, \emph{cabinet}) with an AIRBOT Play arm and a single Azure Kinect DK environment camera, each run from a locked and an unlocked start for $5$ trials apiece. In every task the hidden internal state is unobservable up front and must be resolved by interaction.

\subsection{Setup}
\label{sec:setup}
\begin{figure*}[t]
    \centering
    \includegraphics[width=0.82\textwidth]{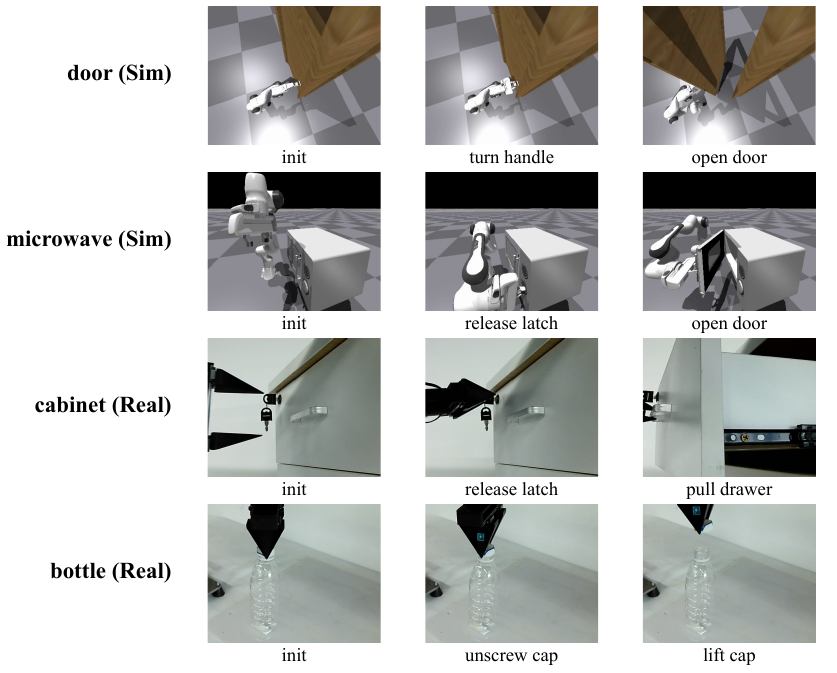}
    \caption{\textbf{Experimental tasks.} We evaluate on four articulated-object tasks with hidden internal state, two in simulation (\emph{door}, \emph{microwave}) and two on a real robot (\emph{bottle}, \emph{cabinet}). For each task we show the initial observation (leftmost) followed by the ordered atomic operations that compose its manipulation procedure (left to right); the hidden state (a latch or lock, which way a handle must turn, or whether a cap is already unscrewed) is unobservable from the initial view and must be resolved by interaction.}
    \label{fig:tasks}
\end{figure*}

\paragraph{Tasks} Figure~\ref{fig:tasks} shows the operations that complete each task; in each, one is a redundant exploratory step that a single uncovering encounter lets IOM remove. Table~\ref{tab:sets} lists the candidate de-redundified procedures $\mathcal{K}$ and the uncovering probe $u \in \mathcal{U}$ for each task.
\paragraph{Policy training} Following AdaManip's training configuration~\cite{wang2025adamanip}, we train a separate policy from scratch for each task on a single RTX 3080, using 20 demonstrations in simulation and 60 on the real robot.

\paragraph{Memory variants} All variants run the same procedure-conditioned policy on the same tasks from the same initial states, differing only in the procedure it receives:
\begin{itemize}
\item \textbf{Random}: a randomly drawn procedure, so the policy re-explores each episode; the no-memory baseline.
\item \textbf{GT} (ground truth): an oracle de-redundified procedure derived from the environment's revealed internal state; an upper bound on what procedure memory can deliver.
\item \textbf{VLM}: the procedure produced by a pretrained vision--language model (GPT-5.5), with no task-specific training.
\end{itemize}
Any difference between them therefore isolates the effect of the stored procedure.

\paragraph{Metrics} We report the number of atomic manipulation operations per successful episode (operation count, $\downarrow$) as the efficiency metric, and the task success rate ($\uparrow$) as a non-regression guardrail; timeouts count as failures.

\subsection{Results}
\label{sec:results}

\begin{table}[t]
\centering
\setlength{\tabcolsep}{4pt}
\caption{Operation count per successful episode on repeat encounters, change versus the no-memory baseline (\textbf{Random}), and success rate, in simulation and on a real robot. The oracle procedure memory (\textbf{GT}) cuts operations by 16--30\% without regressing success; the training-free vision--language memory (\textbf{VLM}) recovers most of that saving with no task-specific training.}
\label{tab:main}
\begin{tabular}{llccc}
\toprule
Task & Memory & ops $\downarrow$ & $\Delta$ vs.\ Random & success $\uparrow$ \\
\midrule
\multicolumn{5}{@{}l}{\textit{Simulation}} \\
\multirow{3}{*}{microwave} & Random & 1.96 & -- & 1.000 \\
                           & \textbf{GT} & \textbf{1.37} & \textbf{$-30.2\%$} & 0.980 \\
                           & VLM & 1.44 & $-26.5\%$ & 1.000 \\
\addlinespace
\multirow{3}{*}{door}      & Random & 2.54 & -- & 0.975 \\
                           & \textbf{GT} & \textbf{2.13} & \textbf{$-16.2\%$} & 0.975 \\
                           & VLM & 2.26 & $-11.1\%$ & 0.975 \\
\midrule
\multicolumn{5}{@{}l}{\textit{Real robot}} \\
\multirow{3}{*}{bottle}    & Random & 2.12 & -- & 0.80 \\
                           & \textbf{GT} & \textbf{1.56} & \textbf{$-26.8\%$} & 0.90 \\
                           & VLM & 1.67 & $-21.6\%$ & 0.90 \\
\addlinespace
\multirow{3}{*}{cabinet}   & Random & 1.86 & -- & 0.70 \\
                           & \textbf{GT} & \textbf{1.56} & \textbf{$-16.2\%$} & 0.90 \\
                           & VLM & 1.62 & $-12.5\%$ & 0.80 \\
\bottomrule
\end{tabular}
\end{table}

Table~\ref{tab:main} reports operation counts and success rates for every variant and task. The deployed system, VLM memory on the shared policy, cuts operations on all four tasks at non-regressing success; the ablations below attribute this gain to the framework and to its training-free instantiation in turn.

\paragraph{A procedure memory amortizes exploration (framework)} The oracle procedure memory cuts operations on all four tasks by 16--30\% relative to re-exploration (Table~\ref{tab:main}). Success does not pay for this: it stays within two points in simulation and, on the real robot (10 trials per condition), the memory variants reach higher success than the no-memory baseline; the real-robot savings are if anything conservative, since Random's failures fall on the harder locked instances and are excluded from its mean. The saving is mechanistic rather than a success artifact, as memory shifts the executed-procedure distribution off the redundant paths: on microwave, the precautionary two-step routine that re-exploration runs in 80\% of episodes is replaced by the direct one-step procedure in 70\% of memory episodes. The gain is precisely the elimination of redundant probes that an abstracted procedure can express but a replayed trajectory cannot.

\paragraph{The training-free instantiation captures most of the gain (instantiation)} Conditioning on the VLM's procedure recovers 69--88\% of the oracle reduction across the four tasks, out of the box and with no task-specific training. We attribute the residual gap to the VLM's accuracy: it is correct on $\approx$88\% of door instances, so the gap is headroom to close, not a ceiling of the framework.

\paragraph{A wrong procedure is recovered from, not obeyed (robustness)} Because the VLM is imperfect, $\approx$12\% of door instances receive a wrong procedure, yet success is unchanged, identical to both GT and re-exploration. The wrong procedures surface only as operations, not failures: the GT--VLM gap is exactly the bounded cost of recovering from them, showing that conditioning as a soft bias on a feedback-driven policy degrades gracefully.

\paragraph{The saving concentrates on repeat encounters} Excluding the first, memory-free episode, in which every variant is forced to explore and the gap is diluted, the oracle's per-repeat reduction in simulation rises to 20--33\%: the amortization is realized on exactly the encounters the method targets.

\paragraph{Recall in place of fine perception} On the bottle, whether the cap is already unscrewed shows only as a narrow gap, too fine to read reliably from images. By resolving the state through action on the first encounter and recalling it by the instance key thereafter, IOM lets later encounters skip this discrimination, turning a brittle perceptual judgment into a lookup, a benefit the operation count does not capture.

\paragraph{Summary} A de-redundified procedure distilled from a single uncovering encounter amortizes 16--30\% of manipulation operations across repeats at no cost to success, in simulation and on a real robot, and an off-the-shelf vision--language model realizes most of this saving with no task-specific training.

\section{Conclusion}
\label{sec:conclusion}
We introduced Instance-Oriented Memory (IOM), a modular, object-centric framework that amortizes manipulation exploration across repeated encounters with an object instance by recording, from a single uncovering encounter, the procedure for manipulating that instance and recalling it instead of re-exploring. Three interchangeable modules realize this: procedure distillation, instance-keyed memory, and a procedure-conditioned policy. Instantiated with off-the-shelf components and no task-specific training, IOM reduces manipulation operations by 16--30\% on repeat encounters across four articulated-object tasks, in simulation and on a real robot, at non-regressing success, with a training-free vision--language model capturing most of this saving.

IOM keys its memory on an instance's identifiable features $\phi(x)$, which it assumes distinguish instances that demand different procedures; when objects look alike yet must be manipulated differently, this key is ambiguous and memory may recall the wrong procedure, the perceptual-aliasing failure mode studied in~\cite{guo2026chameleon}. Storing several procedures under one key and selecting among them at runtime would only defer this ambiguity rather than resolve it. The principled remedy, which we leave to future work, is a more discriminative key that widens $\phi(x)$ beyond appearance to the broader context an instance sits in, such as its surrounding scene, so that objects requiring different procedures are keyed apart.

\section*{Acknowledgments}
The authors acknowledge the use of coding agents (e.g., Claude Code) for assisting in prototype code generation, paper illustration, and improving the clarity of the manuscript.

\balance
\bibliographystyle{IEEEtran}
\bibliography{references}

\begin{thebibliography}{10}
\providecommand{\url}[1]{#1}
\csname url@samestyle\endcsname
\providecommand{\newblock}{\relax}
\providecommand{\bibinfo}[2]{#2}
\providecommand{\BIBentrySTDinterwordspacing}{\spaceskip=0pt\relax}
\providecommand{\BIBentryALTinterwordstretchfactor}{4}
\providecommand{\BIBentryALTinterwordspacing}{\spaceskip=\fontdimen2\font plus
\BIBentryALTinterwordstretchfactor\fontdimen3\font minus
  \fontdimen4\font\relax}
\providecommand{\BIBforeignlanguage}[2]{{%
\expandafter\ifx\csname l@#1\endcsname\relax
\typeout{** WARNING: IEEEtran.bst: No hyphenation pattern has been}%
\typeout{** loaded for the language `#1'. Using the pattern for}%
\typeout{** the default language instead.}%
\else
\language=\csname l@#1\endcsname
\fi
#2}}
\providecommand{\BIBdecl}{\relax}
\BIBdecl

\bibitem{kwon2025rtcache}
O.~Kwon, A.~George, A.~Bartsch, and A.~Barati~Farimani, ``Rt-cache:
  Training-free retrieval for real-time manipulation,'' in \emph{IEEE-RAS Intl.
  Conf. on Humanoid Robots (Humanoids)}, 2025, arXiv:2505.09040.

\bibitem{zhao2026retrievethensteer}
J.~Zhao, H.~Yang, Y.~Hu, Y.~Gao, Q.~Ou, C.~Wan, S.~Dong, Z.~Ma, and Y.~Gong,
  ``Retrieve-then-steer: Online success memory for test-time adaptation of
  generative vlas,'' \emph{arXiv preprint arXiv:2605.10094}, 2026.

\bibitem{wang2025adamanip}
Y.~Wang \emph{et~al.}, ``Adamanip: Adaptive articulated object manipulation
  environments and policy learning,'' in \emph{International Conference on
  Learning Representations (ICLR)}, 2025, arXiv:2502.11124.

\bibitem{wang2022adaafford}
------, ``Adaafford: Learning to adapt manipulation affordance for 3d
  articulated objects via few-shot interactions,'' in \emph{European Conference
  on Computer Vision (ECCV)}, 2022, arXiv:2112.00246.

\bibitem{li2025mapvla}
R.~Li \emph{et~al.}, ``Map-vla: Memory-augmented prompting for
  vision-language-action model in robotic manipulation,'' 2025,
  arXiv:2511.09516.

\bibitem{lee2025vap}
S.~Lee, S.~Mo, and W.-S. Han, ``Bring my cup! personalizing
  vision-language-action models with visual attentive prompting,'' 2025,
  arXiv:2512.20014.

\bibitem{sridhar2026memer}
A.~Sridhar, J.~Pan, S.~Sharma, and C.~Finn, ``Memer: Scaling up memory for
  robot control via experience retrieval,'' 2026, arXiv:2510.20328.

\bibitem{lu2021objectreuse}
S.~Lu, R.~Wang, Y.~Miao, C.~Mitash, and K.~Bekris, ``Online object model
  reconstruction and reuse for lifelong improvement of robot manipulation,''
  2021, arXiv:2109.13910.

\bibitem{zhang2025adaptiveonthefly}
X.~Zhang \emph{et~al.}, ``Adaptive articulated object manipulation on the fly
  with foundation model reasoning and part grounding,'' 2025, arXiv:2507.18276.

\bibitem{wu2022vatmart}
R.~Wu \emph{et~al.}, ``Vat-mart: Learning visual action trajectory proposals
  for manipulating 3d articulated objects,'' in \emph{International Conference
  on Learning Representations (ICLR)}, 2022, arXiv:2106.14440.

\bibitem{wang2026vqmem}
H.~Wang \emph{et~al.}, ``Beyond short-horizon: Vq-memory for robust
  long-horizon manipulation in non-markovian simulation benchmarks,'' 2026,
  arXiv:2603.09513.

\bibitem{memmel2025strap}
M.~Memmel, J.~Berg, B.~Chen, A.~Gupta, and J.~Francis, ``Strap: Robot
  sub-trajectory retrieval for augmented policy learning,'' in
  \emph{International Conference on Learning Representations (ICLR)}, 2025,
  arXiv:2412.15182.

\bibitem{nasiriany2022sailor}
S.~Nasiriany, T.~Gao, A.~Mandlekar, and Y.~Zhu, ``Learning and retrieval from
  prior data for skill-based imitation learning,'' in \emph{Conference on Robot
  Learning (CoRL)}, 2022, arXiv:2210.11435.

\bibitem{dipalo2024kat}
N.~Di~Palo and E.~Johns, ``Keypoint action tokens enable in-context imitation
  learning in robotics,'' in \emph{Robotics: Science and Systems (RSS)}, 2024,
  arXiv:2403.19578.

\bibitem{fu2024icrt}
L.~Fu \emph{et~al.}, ``In-context imitation learning via next-token
  prediction,'' 2024, arXiv:2408.15980.

\bibitem{vosylius2025instant}
V.~Vosylius and E.~Johns, ``Instant policy: In-context imitation learning via
  graph diffusion,'' in \emph{International Conference on Learning
  Representations (ICLR)}, 2025, arXiv:2411.12633.

\bibitem{shinn2023reflexion}
N.~Shinn, F.~Cassano, E.~Berman, A.~Gopinath, K.~Narasimhan, and S.~Yao,
  ``Reflexion: Language agents with verbal reinforcement learning,'' in
  \emph{Conference on Neural Information Processing Systems (NeurIPS)}, 2023,
  arXiv:2303.11366.

\bibitem{zhao2024expel}
A.~Zhao, D.~Huang, Q.~Xu, M.~Lin, Y.-J. Liu, and G.~Huang, ``Expel: Llm agents
  are experiential learners,'' in \emph{AAAI Conference on Artificial
  Intelligence (AAAI)}, 2024, arXiv:2308.10144.

\bibitem{park2023generative}
J.~S. Park, J.~C. O'Brien, C.~J. Cai, M.~R. Morris, P.~Liang, and M.~S.
  Bernstein, ``Generative agents: Interactive simulacra of human behavior,'' in
  \emph{ACM Symposium on User Interface Software and Technology (UIST)}, 2023,
  arXiv:2304.03442.

\bibitem{allard2026erl}
M.-A. Allard, A.~Teinturier, V.~Xing, and G.~Viaud, ``Experiential reflective
  learning for self-improving llm agents,'' 2026, arXiv:2603.24639.

\bibitem{pink2025episodic}
M.~Pink \emph{et~al.}, ``Position: Episodic memory is the missing piece for
  long-term llm agents,'' 2025, arXiv:2502.06975.

\bibitem{zha2023droc}
L.~Zha \emph{et~al.}, ``Distilling and retrieving generalizable knowledge for
  robot manipulation via language corrections,'' 2023, arXiv:2311.10678.

\bibitem{tziafas2024lrll}
G.~Tziafas and H.~Kasaei, ``Lifelong robot library learning: Bootstrapping
  composable and generalizable skills for embodied control with language
  models,'' 2024, arXiv:2406.18746.

\bibitem{sarch2023helper}
G.~Sarch, Y.~Wu, M.~J. Tarr, and K.~Fragkiadaki, ``Open-ended instructable
  embodied agents with memory-augmented large language models,'' in
  \emph{IEEE/CVF International Conference on Computer Vision (ICCV)}, 2023,
  arXiv:2310.15127.

\bibitem{wang2023voyager}
G.~Wang \emph{et~al.}, ``Voyager: An open-ended embodied agent with large
  language models,'' 2023, arXiv:2305.16291.

\bibitem{liu2023reflect}
Z.~Liu, A.~Bahety, and S.~Song, ``Reflect: Summarizing robot experiences for
  failure explanation and correction,'' in \emph{Conference on Robot Learning
  (CoRL)}, 2023, arXiv:2306.15724.

\bibitem{shi2025memoryvla}
H.~Shi \emph{et~al.}, ``Memoryvla: Perceptual-cognitive memory in
  vision-language-action models for robotic manipulation,'' 2025,
  arXiv:2508.19236.

\bibitem{yang2026eventvla}
G.~Yang \emph{et~al.}, ``Eventvla: Event-driven visual evidence memory for
  long-horizon vision-language-action policies,'' 2026, arXiv:2606.20092.

\bibitem{torne2026mem}
M.~Torne \emph{et~al.}, ``Mem: Multi-scale embodied memory for vision language
  action models,'' 2026, arXiv:2603.03596.

\bibitem{bousmalis2024robocat}
K.~Bousmalis \emph{et~al.}, ``Robocat: A self-improving generalist agent for
  robotic manipulation,'' 2024, arXiv:2306.11706.

\bibitem{sharma2023medal}
A.~Sharma, A.~M. Ahmed, R.~Ahmad, and C.~Finn, ``Self-improving robots:
  End-to-end autonomous visuomotor reinforcement learning,'' in \emph{Robotics:
  Science and Systems (RSS)}, 2023, arXiv:2303.01488.

\bibitem{lei2026robomemarena}
H.~Lei \emph{et~al.}, ``Robomemarena: A comprehensive and challenging robotic
  memory benchmark,'' 2026, arXiv:2605.10921.

\bibitem{liang2026metamemory}
B.~Liang, C.~Ke, R.~Zhao, Q.~Zhu, L.~Gui, Y.~Yu, H.~Wang, R.~Xu, and K.-F.
  Wong, ``Meta-memory for large language models,'' \emph{IEEE Transactions on
  Audio, Speech and Language Processing}, vol.~34, pp. 2774--2787, 2026.

\bibitem{zhao2023act}
T.~Z. Zhao, V.~Kumar, S.~Levine, and C.~Finn, ``Learning fine-grained bimanual
  manipulation with low-cost hardware,'' in \emph{Robotics: Science and Systems
  (RSS)}, 2023, arXiv:2304.13705.

\bibitem{wang2023m3e}
Y.~Wang, Q.~Sun, and S.~He, ``M3e: Moka massive mixed embedding model,''
  \url{https://huggingface.co/moka-ai/m3e-small}, 2023.

\bibitem{lin2026la4vla}
T.~Lin, Y.~Du, Y.~Mao, Z.~Ye, Y.~Zhong, B.~Cheng, Y.~Wang, J.~Liu, Y.~Tian,
  J.~Yan \emph{et~al.}, ``{LA4VLA}: Learning to act without seeing via
  language-action pretraining,'' 2026,
  \href{https://arxiv.org/abs/2606.27295}{arXiv:2606.27295}.

\bibitem{guo2026chameleon}
X.~Guo \emph{et~al.}, ``Chameleon: Control-indexed prospective memory for
  visuomotor manipulation,'' 2026, arXiv:2603.24576.

\end{thebibliography}

\end{document}